# AI for the Public Sector: Opportunities and challenges of cross-sector collaboration

## Slava Jankin Mikhaylov[1], Marc Esteve[2,3], Averill Campion[3]

[1]*Institute for Analytics and Data Science, School of Computer Science and Electronic Engineering and Department of Government, University of Essex*
[2]*School of Public Policy, University College London*
[3]*Department of Strategy and General Management, ESADE, Ramon Llull University*

**Keywords:** cross-sector collaboration, data science, artificial intelligence, public policy

## Summary

Public sector organisations are increasingly interested in using data science and artificial intelligence capabilities to deliver policy and generate efficiencies in high uncertainty environments. The long-term success of data science and AI in the public sector relies on effectively embedding it into delivery solutions for policy implementation. However, governments cannot do this integration of AI into public service delivery on their own. The UK Government Industrial Strategy is clear that delivering on the AI grand challenge requires collaboration between universities and public and private sectors. This cross-sectoral collaborative approach is the norm in applied AI centres of excellence around the world. Despite their popularity, cross-sector collaborations entail serious management challenges that hinder their success. In this article we discuss the opportunities and challenges from AI for public sector. Finally, we propose a series of strategies to successfully manage these cross-sectoral collaborations.

*Slava Jankin Mikhaylov (s.mikhaylov@essex.ac.uk).

† Institute for Analytics and Data Science, University of Essex, Knowledge Gateway, Colchester, CO4 3AD UK

# Introduction

An ambition to be the world's most innovative economy is set out in the UK Government Industrial Strategy. Artificial Intelligence (AI) Sector Deal institutionalises the partnership between government, industry, and academia in achieving this key ambition by aiming to attract and retain domestic and international AI talent; deliver upgrades to digital and data infrastructure; ensure business climate conducive to starting and growing an AI business; and contribute to prosperity of society by spreading AI benefits across the country [1]. The AI Sector Deal outlines the commitment from government, industry, and academia as the three partners with a sectoral support package of around £1bn that complements additional £1.7bn under the Industrial Strategy Challenge Fund [1].

This cross-sectoral partnership is also built in the governance structure of the AI Sector Deal [1]. Oversight of the implementation of the Deal and maximisation of its potential will be led by the new government Office for AI. The office will support the AI Council that will bring together leaders from industry and academia to provide strategic leadership and drive the implementation of the Deal. Societal benefits of AI will be ensured through the creation of a Centre for Data Ethics and Innovation to advise on the ethical use of data and AI.

For the future industry of the UK, AI sector has the potential to provide over 80,000 new jobs [1: p.36], adding £200bn or 10 per cent of UK GDP by 2030 [1: p.24]. AI also holds significant promise for public sector that is undergoing transformation with robotics and automation changing the provision of public services [2]. The challenge for AI adoption in public sector is to better use citizen data for improvement of public services. The direct value of citizen data held in public sector has been estimated at £1.8bn, with wider social and economic benefits totalling £6.8bn [3]. These data can be used to more specifically target who needs public services and 'tailor those services more accurately' [2].

Governments have historically taken on the role of an entrepreneurial state playing a significant role in innovation [4]. A well-known example of such initiatives is the Defence Advanced Research Projects Agency (DARPA) of the US, which has developed the technology behind the internet and personal computer [5]. The UK has also seen multiple examples of how this can be implemented in practice, with initiatives such as Innovate UK, the Small Business Research Initiative (SRBI), and The Catapult Programme, among others.

The ambition of the AI Sector Deal is amplified by external events like the Brexit, but ultimately depends on successful collaboration among the three partners. Such collaboration has become a fundamental activity in most, if not all, entrepreneurial state initiatives [6]. Public organisation adoption of AI and data science presents numerous known challenges ranging from employee path dependency on embedded processes and norms, information silos, a lack of resources, collaborative culture, and technical capacities [7, 8]. Successful delivery of the AI Sector Deal relies on collaboration across three partners. It is, therefore, important to understand challenges and success factors for such cross-sectoral collaboration learning from collaborative experiences in other sectors. The present study synthesises existing knowledge on how to manage cross-sector collaborations and proposes a series of recommendations on how they should be considered to integrate AI and data science initiatives into public service delivery.

## Mapping Inter-Organisational Collaboration in AI

Delivery of public services is often implemented via different types of organisational forms bringing together public, private, and non-profit actors [9]. These collaborations may imply highly formalised ventures, such as public-private partnerships, or more informal arrangements, such as policy networks. Their main rationale is to merge the strengths of each involved party to increase the effectiveness and the value for money of a particular public service. Cross-sector collaboration has gained importance worldwide, particularly across European countries [10]. In the UK, for instance, more than 725 public-private partnerships worth over £54.2bn have been developed to create hospitals, schools, prisons, bridges, roads, and military equipment [11]. Despite their popularity, collaborations entail numerous management complexities [12, 13] and as a result, a high percentage of them frequently do not achieve satisfactory outcomes [14], while some are highly successful [15].

There are existing tri-lateral collaborative arrangements in the area of AI and data science that provide the knowledge base. For example, the University of Essex arranged a joint appointment with Essex County Council of a professorship in the area of public policy and data science as Chief Scientific Adviser to the Council that is based in a specially designated institutional vehicle -- the Institute for Analytics and Data Science (IADS) within School of Computer Science and Electronic Engineering. IADS is a centre of excellence at the University of Essex that connects scholars, businesses, institutions, and government for the AI related work. The aim of the relationship is to lever resources and data of the public sector with AI expertise of the University and businesses to deliver services for the benefit of community in Essex. Table 1 below provides an overview and summary of several additional examples of similar collaborations from around the world.

**TABLE 1**

*Cross-sectoral AI and Data Science Centres of Excellence*

| Centre | Participants | Participant Type | Aims | Core areas of application |
|---|---|---|---|---|
| Institute for Analytics and Data Science (IADS)<br><br>Location: UK | University of Essex, Essex County Council, Suffolk County Council, British Telecom, EPUT NHS, UNESCO | University; Local government; NGO; organisation; Private utilities | To create new products and services for businesses, individuals, and society; to facilitate knowledge transfer around AI between academia and private, public and third sectors | International development, public policy, healthcare, social care, mental health, insurance, finance, telecoms, transport, media, policing and crime prevention |
| Singapore Data Science Consortium (SDSC)<br><br>Location: Singapore | National University of Singapore, Nanyang Technological University, the Singapore Management University, Agency for Science, Technology and Research, National Research Foundation Prime Minister's Office, Defence Science & Technology Agency, Singapore Tourism Board, ST Electronics, GIC, Micron, Fuji Xerox, Surbana Jurong, Certis Cisco, ASM Assembly Systems, Television Content Analytics TVCONAL | Universities; National research agencies; Private tech companies; Local government | To facilitate collaboration between institutes of higher learning, research, industry and government in data science R&D | Healthcare, customers and retail, manufacturing, transport |
| AI Singapore<br><br>Location: Singapore | National University of Singapore, Singapore University of Technology and Design, Nanyang Technological University, Agency for Science, | Universities; National research agencies | To catalyse, synergise, and boost Singapore's AI capabilities, to use AI to address major challenges that affect | Healthcare, urban mobility, cybersecurity, computing platform, privacy preserving |

| | Technology and Research, Singapore Management University | | society, and to invest in deep capabilities to catch the next wave of innovation | technologies, sensing and measurements |
|---|---|---|---|---|
| Beijing Institute of Big Data Research (BIBDR)<br><br>Location: China | Peking University, Beijing University of Technology, Zhongguancun Science Park, Haidian District government under supervision of municipal government of Beijing | Universities; District government | To combine education, research, entrepreneurship, and government service to create world class program for developing data science in China and a platform for nurturing new enterprises in big data | Healthcare, traffic, finance |
| RMIT Data Analytics Lab<br><br>Location: Australia | RMIT University Melbourne, NICTA (NSW government, Queensland government), Australian Research Council | Universities; Regional government; National research council | To become a hub for advanced data analytics projects to help Australian business compete on a global scale | Geospatial information search, biomedical informatics for health decision making, integrated design infrastructure for Australian cities |
| The GovLab-NYU<br><br>Location: USA | NYU Tandon School of Engineering, White House Office of Science and Technology, Laura and John Arnold Foundation, MacArthur Foundation, The Australian National Government, England National Health Service, | Universities; National research institute; Private foundation; Foreign government partners; NGOs | To strengthen the ability of institutions and people to work more openly, collaboratively, effectively, and legitimately to make better decisions and solve public | Criminal justice, healthcare, government innovation, public decision making |

| | UNICEF, Omidyar Network | | problems with big data and open data | |
|---|---|---|---|---|
| California Policy Lab<br><br>Location: USA | UCLA, UC Berkeley, Californian governments local, county, and state levels | Universities; State departments; County government; Local government | To create data driven, scientific evidence and insights to help government at all levels in the state solve urgent problems; to help bridge the gap between policy makers in the research community | Homelessness, poverty, crime, education inequality |
| Center for Data Science and Public Policy<br><br>Location: USA | University of Chicago Harris School of Public Policy, Computation Institute, Municipality of Rotterdam, Charlotte-Mecklenburg Police Department, Metropolitan Nashville Police Department, San Francisco Police Department, Los Angeles Sherriff's Department, Chicago Department of Public Health, Chicago Department of Innovation and Technology, Environmental Protection Agency | Universities; Local government; County government; Research institute; Public research; National government | To educate current and future policy makers, doing data science projects with government, non-profit, academic and foundation partners, and developing methods and open source tools that support and extend use of data science for public policy and social impact | Welfare, city infrastructure, citizen engagement, highway patrol, urban planning |
| Dalle Molle Institute for Artificial Intelligence (IDSIA) | Swiss Confederation Commission for Technology and Innovation, University of Lugano, University of | Universities National research institute; | To offer solutions to a range of complex problems through theoretical findings and novel algorithms, | Military decision making, metrology and climatology, environmental risk |
| Location: Switzerland | Applied Sciences and Arts of Southern Switzerland, Imprecise Probability Group (IPG), Swiss National Science Foundation, Federal Department of Defence | National government; Research network | machine learning, deep neural networks, and imprecise probabilities by promoting strong cooperation with partners | analysis, bioinformatics |
| German Research Centre for Artificial Intelligence (DFKI) Location: Germany | University of Bremen, Deutsche Forschungsgemeinschaft, Deutschland Land der Ideen, Berlin Big Data Center | Universities; National government; Research institute | To study design, realisation, and analysis of information processing models that enable robotic agents and humans to master complex human scale manipulation tasks that are mundane and routine | Emergency response and crisis management, outreach, multimedia opinion mining |
| Insight Centre for Data Analytics Location: Ireland | Dublin City University, NUI Galway, University College Cork, University College Dublin, Cisco, Intel Corporation, Tyndall National Institute, HP, Central Statistics Office, Open Data Institute, Dublin City Council, Galway City Council, Department of Public Expenditure and Reform | Universities; Councils; National government; Research institute; Private sector | To use information to make decisions based on it for transformation by taking the guesswork out of decision making in society | Personalised public services, chronic disease management and rehabilitation, smart enterprise, open government, urban life quality |
| EBTIC Location: UAE | Khalifia University-Abu Dhabi campus, ICT fund-Telecommunications Regulatory Authority, Etisalat, BT | Universities Private utilities; National government | To collaborate with industry, universities, and government organisations to be a driving force for | Smart infrastructure, smart network design, smart |

| | | | innovation for the Middle East region | society, smart enterprise |
|---|---|---|---|---|
| | | | | |

Table 1 illustrates that across continents, governments are engaging with universities and a variety of sectors through policy lab platforms in order to combine different capabilities and problem solving capacities. These cross-sectoral labs help synergise knowledge so that government can work towards AI and data science based solutions in areas ranging from prevision medicine to smart cities. For instance, The GovLab at NYU frequently develops applied research frameworks that help government approach problems in a more data-informed, innovative manner. The Lab's People Led Innovation Methodology is used by city officials to approach major public problems through a series of four phases that unleash the expertise of others to create solutions. These AI and data science labs often provide a network for training and skill development for public servants and recommend communication and technical advice for 'smarter' management.

The benefits of adopting AI drive the incentives for collaborative arrangements. These benefits relate to prediction and anticipation of demand for services, automation of demand-side response, identification high-risk groups and development of targeted interventions; production of goods with higher productivity, lower cost, and better efficiency; promote products and services at the right price, with the right message, and to the right targets; and provide enriched, tailored, and convenient customer experience [16].

Social benefits of such collaborations focus around improving public service delivery and relieving administrative burdens. For example, Essex County Council predict the risk of 14 year olds becoming NEET (not in education, employment or training) by age 18 and work with schools to develop early-stage interventions with additional support to encourage those at high risk of becoming NEET to remain in employment or education [3: p.37]. The UK Cabinet Office Behavioural Insights Team showed how to analyse initial referral and assessment notes in social care to predict closed case escalation (how many cases would come back into the social care system) [17]. Through overcoming resource constraints, paperwork, and backlogs in a more cost-efficient, effective, and time-savings manner, government gains the opportunity to exist as an empathetic service provider [18]. To illustrate, DARPA's 'Education Dominance' program uses AI to reduce the time required for Navy recruits to become technical experts (from years to months) through the creation of a digital tutor that applies machine learning to model novice-expert interaction [19]. This resulted in the program recruits outperforming experts with 7-10 years of experience and ensured the recruits' likelihood of securing high tech jobs with high incomes.

Cross-sectoral collaborative efforts around AI are often institutionally organised around offices for data analytics (ODAs). Table 2 below provides a broad overview and summary of major international efforts in this area.

**TABLE 2** *Cross-sectoral Data Analytic Centres*

| Office of Data Analytics | Participants | Participants | Aims | Core areas of application |
|---|---|---|---|---|
| City of Boston Analytics Team<br><br>Location: USA | EMS Boston, Office of New Urban Mechanics, Boston Fire Department, Inspectional Services Department, 311 Call Center, Data for Democracy | Local government; Nonprofit | The aim is to act as a central data organisation interested in using data and maps to create a better understanding of Boston and to use data to improve city public policies | Geospatial, city services, permitting, environment, transportation, emergency response, citizen engagement |
| DataLA<br><br>Location: USA | Ash Center for Democratic Governance and Innovation at Harvard Kennedy, USC Spatial Science Institute, UChicago CDSPP, UCLA, City Parking, County Department of Public Health, County of Los Angeles Bureau of Land Management, LA Sanitation Department | Universities; Regional government; Local government | To work with academics, city departments, the community, sister cities, and private partners to develop insights and digital tools that make the City more liveable and equitable by sharing data | Publishing and maintaining Open Data, geographic data and communication, Open Budget LA, street cleaning efficiency |
| MODA NYC Analytics<br><br>Location: USA | Department of Information Technology and Telecommunications and Technology, NYU Center for Urban Science and Progress, Columbia University Institute for Data Sciences and Engineering, Department of Citywide Administrative Services | Universities; Local government; National government | The aim is to partner with agencies to create, test, and improve analytic models that deliver measureable value to City services and to serve as NYC's civic intelligence center | Crime, public safety, quality of life issues, city-wide data sharing platforms, training and skill development for city employees |
| DataSF<br><br>Location:<br>USA | Open Data Services Team from Department of Technology, Department of Health, San Francisco Department of Environment, San Francisco Art Commission, Code for America | Local government; State government; Nonprofit | The aim is to empower the use of data across City departments so that evidence based policy making and operational improvements can be made | Cross-departmental open data sharing, data governance, data quality standard setting, mother and child nutrition, training and skill development for city and county staff, city greening |
| SmartDubai<br><br>Location:<br>UAE | Mohammed Bin Rashid School of Government, Dubai Health Authority, Dubai Police, Roads and Transport Authority, Department of Economic Development | University; Local government; National government | The aim is to collaborate with private sector and government partners to empower, deliver, and promote efficient, safe, and impactful city experiences for residents and visitors | Data leadership skill development and training, mapping city infrastructure, e-services, environment, mobility |
| Office of Open Data and Digital Transformation-Philadelphia<br><br>Location:<br>USA | University of the Arts Design for Social Impact Program, Division of Housing and Community Development, Department of Planning and Development, Penn Medicine's Center for Health Care Innovation | University; Local government; State government | The aim is to create digital services that support the success and well-being of all Philadelphians to empower them through dignified, accessible and efficient services | Public open data, human catered service design, data sharing platforms, citizen engagement, housing accessibility, historic site vulnerability, public health |
| Greater Manchester Connect<br><br>Location:<br>UK | University Hospital Morecombe and Cumbria Information, Transport for Greater Manchester, Manchester City Council, Health Innovation Manchester | Regional government; Local government | The aim is to put Greater Manchester at the forefront of data-sharing and analysis to help improve public services by establishing a data sharing authority to break down the barriers which stop public services from sharing information. | Health and social care and wider reforms of public services, information governance platforms, data sharing engines, employment and skills, housing, transport |

| | GDS, Department for Communities and Local Government | National government; Local government | The aim is to bring together investment in research, data and intelligence to support the delivery of the region's Strategic Economic Plan, and provide an evidence base for future changes in public services | Open data |
|---|---|---|---|---|
| ODA for the West Midlands<br><br>Location:<br>UK | GDS, Department for Communities and Local Government | National government; Local government | The aim is to bring together investment in research, data and intelligence to support the delivery of the region's Strategic Economic Plan, and provide an evidence base for future changes in public services | Open data |
| Government of South Australia ODA<br><br>Location:<br>Australia | Australia Bureau of Statistics, Department of the Premier and Cabinet, Australia Renewable Energy Agency, Clean Energy Council, Bureau of Meteorology | National government | The aim is the provide high quality data analysis and to support South Australia government agencies and better the State of Australia | Child protection, gender equality, domestic violence, energy |
| SmartDublin<br><br>Location:<br>Ireland | Dublin City Council, South Dublin City Council, Fingal County Council, Comhairle County Council, Intel, IBM, Maynooth University, Lero, Insight | Local government; Regional government; Private tech companies; Universities | The aim is to create a mix of data-driven, networked infrastructure, fostering sustainable economic growth and entrepreneurship, and citizen centric initiatives | Energy monitoring, public transportation passenger information, civic engagement and citizen empowerment, dashboards, trash, traffic |

ODAs are commonly used as a collaborative organisational form for sharing data to improve city services. What differentiates ODAs from policy labs is that policy labs operate as a networked platform often rooted in universities, whereas ODAs are physical offices often associated with the mayor or city manager's office. One example is the Mayor's Office of Data Analytics in New York City (MODA NYC) which serves as a unifying space for aggregating data from across City agencies to more effectively address crime, public safety, and quality of life issues. One contribution from MODA NYC is the creation of a Citywide data sharing platform, Databridge, which 'combines automated data feeds from 50 plus source systems across 20 agencies and external organisations to warehouse and merge geographic information to enable cross-agency analysis' [20]. While ODAs commonly work across agencies within government, Table 2 portrays how other organisations often contribute to knowledge sharing and analysis in these settings. In order to develop standards and protocols for data sharing, MODA NYC actively partners with the expertise of NYU Center for Urban Science and Progress as well as Columbia University Institute for Data Science and Engineering. Thus, ODAs also take on a 'data liaison' role as a 'designated point of contact for outside partners contributing to or using City data' [20].

Such cross-sectoral collaboration is as promising as it is challenging. While the benefits of interacting across sectors to implement AI strategies are many, working across sectors has been proven to be very complex [8]. The next section discusses challenges and success factors of cross-sectoral collaboration that should be considered for successful delivery of the AI Sector Deal.

# Challenges and opportunities of cross-sectoral collaboration around AI

In order to analyse the managerial practices influencing collaborative arrangements across different sectors, we used a systematic review of the existing literature. The search strategy used to find eligible studies was carried out with an electronic search in Google scholar database. We chose our publication criteria to be only those public administration and management articles found in, arguably, the top three journals in the field: *Journal of Public Administration and Theory, Public Administration Review, and Public Management Review*. The following search terms were used `collaboration performance', `collaboration success', `network performance', `network success', `joint venture performance', and `joint venture success.' These searches lead to 7,885 results initially; once the journal filter was added 156 results were acquired. Next, after searching the abstract and title for the relevant terms a total of 84 articles were included in the review. We also used a `study design' filter and kept only research with empirical evidence (e.g. articles that use research design such as case studies, surveys, questionnaire) on factors for success and performance in collaboration, networks, and joint ventures. For each article, we summarised the authors, publication year, title, journal, success factors identified, effectiveness determinants identified, and managerial strategies that led to success. By examining only articles with the keywords `success' and `performance' we were able to extract determinants that led to success as identified in the empirical studies. In order to analyse the selected studies, we engaged in an inductive analytic process [21] to derive the main factors influencing success in collaborative ventures.

## Challenges for successful collaboration

While the challenges of collaboration across private sector organisations have been widely researched [22] much less attention has been paid to the difficulties of working across public, private and non-profit sectors. Scholars such as Stoker et al. [23] have focused on the development of social capital as a means to address cross-sectoral collaborative challenges that are rooted in the way actors perceive each other's abilities to relate to one another. Additionally, in the environmental policy field, Innes et al. [24, 25] have contributed insights about the benefits of network structures in overcoming traditional bureaucratic based institutional constraints through their self-organising and adaptive nature.

More recently, Andrews et al. [26] reviewed the empirical works on the differences among public and private organisations and propose a series of arguments on how these differences challenge inter-organisational collaborations, based on mixing environments, structures, goals, and values. The first issue that can hinder collaboration success is the different environments surrounding public and private organisations. While public organisations are accountable to their service users and, also, to the public at large, private organisations respond to their shareholders [27]. This can lead to clashes when aligning the interests of the different partners engaged in the collaboration [26, 28]. In a recent report on AI by the House of Lords [29] new questions of accountability were raised. Public sector procurement of artificial intelligence based technologies presents challenges regarding the 'legal liability where a decision taken by an algorithm has an adverse impact on someone's life' or 'the potential criminal misuse of artificial intelligence and data' [29: p.95]. There is a duality surrounding the positive impacts that data-based decision making tools and machine learning can have on public policy making and implementation. If things go wrong, pinpointing responsibility becomes a web of closely inter-linked realities: 'Is it the person who provided the data? The person who built the AI? The person who validated it? Operates it?' [29: p.309]. Additionally, in the case that the business which creates the technology is responsible, it is not unlikely these companies exist overseas, in places like China or Singapore, potentially turning jurisdictional action into a sand-trap of international law.

Another central aspect of the challenges associated with mixing environments in cross-sectoral collaborations is the divergent approaches to managing risk in the public and private sectors. Klijn and Teisman [30] argue that the political risks of government are not easily reconciled with the market risks of business organisations. As future inter-organisational collaborations related to AI take place, there is always the inherent risk that the data used has been gamed or sabotaged to serve the opportunism of a self-interested actor. For example, this 'needle in the haystack' situation can occur in training or operation phases like the intentional use of misleading data fed into systems or 'destroying, altering, and injecting large quantities of misleading data' [29].

Competing institutional logics have been considered a fundamental challenge when developing collaborative ventures [31]. Public organisations operate in what has been described as a state logic, while private organisations operate with a mix of market-based and corporate logic. These authors expose that business partners in collaborations 'conflate their role as shareholders – thus invoking the market logic – and their experience as businessmen, as they are accustomed to operating under the Corporate logic within their companies' [26: p.347]. In practice, this influences the agreement of which goals need to be pursued by the collaboration. In particular, managers of collaborative ventures may find it difficult to deliver public value for money, and also to maximise profits to satisfy shareholders of the private partners [32]. Another important aspect affecting collaboration success is the mix of different organisational structures. In this sense, public sector organisations are classically identified as rule-orientated because of the need to meet demanding statutory requirements for due process. By contrast, Rainey [33] describes how private firms are not subject to the same kind of political accountability pressures and so are thought to be less hampered by bureaucratic oversight.

Opportunism in strategic collaborations has been linked with divergence among the organisational values of all involved partners [34]. This is of special importance across cross-sector collaborations, as each sector has been related with a subset of different values - i.e. public employees are more motivated to serve the public, while their private counterparts seek to further their organisation's interests [35]. In a nutshell, then, the major challenge caused by the mixing of values in collaborative ventures involving public, private, and non-profit sectors is to help the members of each organisation to switch their mentalities from the 'us and them' to 'we' [36]. For instance, while cross-sector collaboration forms for AI and data science can be commonly based on procurement or contracting out, university-public sector forms present an opportunity to unite organisational differences through mission-oriented projects. The hope is that projects centred on addressing societal problems will align similar values between organisations due to the 'social good' nature of these projects.

Elaborating on this sentiment, Healy [37] argues that the fragmentation of values in collaborative governance can be united when 'substance and process' are recognized as 'co-constituted, not separate spheres' [37: p.112]. The engagement in the governance process 'shapes participants' sense of themselves; and generates ways of thinking and acting that may be carried forward' with the emergence of a social order [37: p.112]. Creating opportunities to overcome the value challenge is a key task for public managers. To illustrate, The White House Police Data Initiate uses inter-organisational collaboration (e.g. engagement with academics, technologists, police departments) to experiment with machine learning techniques that review audio and video footage from body cameras [38]. Here, the data presented from body cameras serves as a vehicle for knowledge sharing amongst different actors like Stanford University academics and the City of Oakland police department over the benefits and problems of this tactic. From this knowledge sharing process and exchange of perspectives, social and relational substance begins to bridge individual and institutional differences towards collective values and action.

Additional challenges related to cross-sectoral collaboration around AI relate to skills and data. There is a significant skills gap in AI between public sector, on the one hand, and businesses and universities on the other hand. AI Sector Deal is providing for significant investment in skills and people for wider UK economy. However, there is less consideration to the AI skills gap in government (and wider public sector). Moreover, Chen et al. [39] associate public organisations as lagging in individuals who possess the 'prerequisite knowledge and skills to be effective participants' in cross-boundary data based initiatives, and thus require technical assistance and training. In particular, these issues relate to a lack of the technical jargon needed to personalize

'disparate public data' from different organization so that it tells a story; user ability to operate new systems; and incentivizing information sharing at the individual level. At the same time, developing the digital skills needed for public sector use of artificial intelligence is not a quick process, and more funding is needed for PhD students in machine learning to overcome this general shortfall [29]. Extending the system of secondments across three partners can be an immediate solution that could also kick start the knowledge transfer. Classically, Weber et al [40: p.335] find that the 'transfer, receipt, and integration of knowledge across participants' is a constant challenge for any public problem being addressed in inter-organisational settings. Hence, managers must strategically encourage employees to share new information and skill development with co-workers 'to enhance the collective improvement of knowledge' [41: p.699].

Open data initiatives have been largely successful in unlocking commercial value from publicly held data. However, some of the most valuable data for AI innovation cannot be openly shared due to commercial sensitivity, security, or personal information. Successful development of the AI sector relies on developing deeper data sharing relationships across three partners, with the barriers ranging from trust and cultural concerns to practical and legal constraints. Poorly implemented data sharing programmes risk derailing innovative AI cross-sectoral collaborations as witnessed from the case of DeepMind and the Royal Free London NHS Foundation Trust [29, 42]. AI Sector Deal aims to address this issue through the establishment of Data Trusts providing clear frameworks for fair, equitable, and secure data sharing [1: p.30]. Without this type of central information system, the technical capacity to share data across inter-organisational forms is hindered by the fragmentation of data standards [39]. The scholars also suggest that in addition to data sharing standards, forming data collection standards and data quality assurances before it is shared amongst organisations further augments the technical capacity for joint action.

The establishment of a Centre for Data Ethics and Innovation will also ensure safe and ethical use of AI, while the General Data Protection Regulation (GDPR) and the UK Data Protection Bill provide legal certainty over the sharing and use of data, and fair and transparent application of AI [1]. Beyond concerns of ethical use of AI and fairness in the sharing and application of AI, unintended consequences also relate to the 'consequential decisions about people' often made by humans that will be replaced by AI and safety as more 'AI [is used] to control physical world equipment [43: p.30]. Specifically, the report raises concerns about how to ensure justice, fairness, and accountability in this decision making and how machine learning systems will react to the 'complexities of the human environment.' Especially in relation to the criminal justice context, institutions such as the Centre for Data Ethics and Innovation must constantly push for the incorporation of data that is as complete and unbiased as possible. Otherwise, we risk 'exacerbat[ing] problems of bias into these new technological interfaces and 'hardwire discrimination'; however, data analytics can also be used to 'predict and detect bias and prevent discrimination.' [44].

Andrews et al. [26] conclude by highlighting the many potential benefits of cross-sector collaborations but warn that the management complexities that they entail hinder, in most cases, the possible benefits of bringing the strength of different sectors and combining them in a particular project. Thus, understanding the success factors related to collaborations across sectors, together with the different managerial approaches to mitigate the abovementioned challenges, becomes of great importance if initiatives related to, for example, AI projects, need to be implemented through public, private, and non-profit collaborations.

## Success factors of collaboration

Research on the performance of collaborations is vast [45, 46, 47]. One of the main concerns within the literature on collaboration performance has to do with the managerial strategies that can increase the performance of these complex organisational arrangements [48]. Among the reviewed studies, seven main managerial strategies can be distinguished: facilitative leadership; shared objectives; knowledge gathering and sharing; communication; socialising; expertise; and sense-making.

*1) Facilitative Leadership:* Opposite to the classic idea of a hierarchical leader that imposes his or her views towards followers by relying on his or her power position within the organisation, facilitative leadership '*endorses respect and positive relationships among team members, constructive conflict resolution, and candid expression of thoughts and attitudes*' [49]. Ansell and Gash's [50] meta-analysis of the literature on the management of collaborative governance concludes that leaders of collaborations should promote broad, active participation; ensure broad influence and control; facilitate productive group dynamics; and extend the scope of the process. Therefore, it is argued facilitative leadership is imperative to collaboration, especially since incentives to participate can be low and resources may often be asymmetrically distributed. Another implication is that the authors implicitly derive that collaboration performance is determined by achieving a cycle of communication, trust, commitment, understanding, and outcomes. The success of collaboration is implicitly contributed to a combination of face-to-face dialogue (although this alone is not sufficient), the ability to establish trust in the various phases from negotiation to implementation, the level of commitment from stakeholders (which requires cooperation and responsibility to results of consensus and 'ownership' of the decision making), as well as a shared understanding of what can be achieved through working together viewed as common ground or common purpose. It should be noted, however, that facilitative leadership style is not the only leadership approach to manage collaborative ventures. In a recent analysis of cross-sectoral collaborations to deliver water services in Norway, Hovik et al. [51] found that the leadership styles can be contingent to the partnership's characteristics, and the characteristics of the manager determine their ability to understand how to leverage different skills needed. These authors identified four main roles of collaborative leaders: the convener who ensures information flow, role clarification, and compliance; the catalyst who creates motivation, raising awareness, and ensuring ownership; the mediator is the broker and facilitates discussion; and the bridge builder links aims at different levels and ensures political anchorage.

*2) Shared Objectives:* The definition of the organisation's objectives has been positively correlated with the organisational performance of the public sector [52]. However, since alliances involve both joint value creation and value appropriation, these mixed motives may create tension between shared and private objectives [53]. The simultaneous pursuit of different objectives leads managers to continue working in the ways they were used to, because they do not know what objectives to focus on [54]. This lends some support to Jensen's [55] assertion that asking managers to pursue multiple objectives creates problems – not '*confusion and lack of purpose*' as he suggests, but rather a '*status quo bias*' [54]. Even if all the parties in a collaboration are highly aligned with the main objective of the alliance, there may be differences between the objectives of each organisation. The importance that objectives have for collaborations is explained because they '*act as a guide for decision making and a reference standard for evaluating success*' [56].

*3) K*nowledge gathering and sharing: To overcome process and dynamics issues of collaborative governance, Chen and Lee [39] suggest management activities should focus on institutional capacity building for joint-action, like the creation of common standards for the collection and processing of data. On a technical level, their in-depth case study finds that the federally mandated metropolitan planning organisations are challenged by the management of the collaborative data networks necessary to create data sharing across jurisdictions, which is required for more integrated metropolitan transportation planning. The main implication is the way in which knowledge is represented across the divisions of functional departments can nonetheless enable or hinder the improvement of cross-boundary data sharing. Therefore, the formulation of common standards for data collection and sharing is best developed by activating key network members, so that groups are aligned based on their functional responsibilities across the network [39]. For example, frontline workers are knowledge-banks that can support and recommend the design of procedural standardisation. The authors found that the GIS members maintain regular contact and communication in regional transportation planning activities and can contribute specialist insights of data and technology operation. To this end, Nesta [57] recommends that in collaborating with the Greater London Authority and data science specialists to develop an algorithm that predicts which of the City's thousands of properties are unlicensed 'House(s) in Multiple Occupation,' the first step was to speak to building inspectors about the features on a typical HMO, due to their frontline knowledge competencies. Furthermore, it is suggested that improved data sharing procedures derived from using that information to then identify the relevant datasets connected with these criteria. Accurately building the institutional and technical capacity to guide the collaboration requires incorporating quality knowledge at the beginning of the process, so that fragmentation won't '*create issues later on for data integration*' [39].

*4) Communication:* In a recent study, Ansell and Gash [58] have described the different effects that a communication strategy can have on the management of a collaboration. First, they describe the attractor effect, which occurs when it appears that the collaboration is producing tangible outcomes, so stakeholders are more willing to invest time, energy and resources. This happens by showing the value of joint-action through quick wins. Positive learning feedback is another determinant of success in that creating 'politically neutral' spaces for joint learning reduces cultural barriers and increases networking [59]. Learning feedback thus occurs when the knowledge gained from learning how to work together is continuously built upon in subsequent interactions. Next, collaborative platforms are successful when they can exercise architectural leverage, which is achieved through developing shared assets, designs, and standards that can be reconfigured, resulting in multiplier effects built on these pre-existing efforts. Similar themes of performance determinants from past literature are also cited, such as a 'champion' to mobilise support and activity coordination.

5) *Socialising:* When managers make the impact of the efforts of the collaboration transparent and enticing for key players to work together, the collaboration will be positively affected [60, 61]. This seems parallel to the beneficiary contact findings that when employees see how their results impact a person, they increase performance. Transparent results and indicators can facilitate more ideas and reforms throughout all levels of the collaboration where it may be more difficult to implement a top down idea in decentralised settings. The U.S. healthcare sector has been experimenting with this more holistic approach to understanding the 'whole' system surrounding problems in their reimbursement payment schemes through the use of big data collaboration [62]. By combing insights of health data from clinicians, researchers, and patients, there is a shift away from 'isolated and potentially uncoordinated instances of treatment- or *fee for service*-[towards] paying on the basis of better health outcomes' [62: p.22]. In contrast to a former fragmented analysis, an interconnected 'learning system' is quicker to transfer knowledge from different levels back to providers. As Page [61] states, '*inclusive processes used to design and deploy the results and indicators help ally partners' mutual suspicions and turf differences at the beginning of the reform process*' (p.333). This democratises the power in the setting in a way, because once it is known where everyone stands, the next phase of the collaboration can use that information to build upon into their practices or efforts and not just keep the results isolated to the reform architects. This is further explained 'because the vision, mission, goals, and daily activities of collaboration transcend particular individuals and organisations, the results and indicators were necessary to foster organisational commitments to common values and practices' [61: p.333]. Thomson and Perry [63] complement these ideas by defending the importance of shared responsibility. Arguably, arriving at the general consensus needed to manage the collaboration requires an equilibrium where conflict can still occur but within a larger framework of a jointly determined agreement about the rules.

6) *Expertise*: Overall, the manager's mind-set determines the choices about when and how to use analytical tools and strategies needed for the transfer, receipt, and integration of knowledge across the network [40]. More generally, hiring tech savvy network managers and shepherding the efforts of field experts within the network can both induce trust based on their competencies, as well as improve the quality of service [64]. From an operational standpoint, Chen [39: p.15] exemplifies that '*the appropriate use of relevant technology can significantly improve performance in data quality, data integration, data analysis, and visualisation.*' In an analysis of best practices surrounding data-based collaborations in the public sector, NYU GovLab [65] suggests that managers should tackle a lack of institutional readiness by '*utilising the host of data legacy managers already working in government*' like GIS teams for their wealth of information and tactical expertise. From piloting the London Office of Data Analytics, Eddie Copeland [57] also highlights in a lecture that managing downwards in public organisations can liberate the already talented data analysts, who are likely '*stuck reporting on* *monthly dashboards and key performance indicators,*' instead of being used in other administrative capacities. Strategies for integrating the correct knowledge for data analysis in public sector collaboration can further be enhanced through engagement with experts in the field. In particular, during a project with the Municipality of Rotterdam's Rijkwaterstaat traffic patrol department, the University of Chicago Centre for Data Science and Public Policy [66] attributes their first-hand observation of how the relationship works between inspectors patrolling the highways, and traffic control centre's reaction, as a means for designing a more successful data-driven approach to the deployment locations for patrolmen. As legitimacy is a key factor in network success, collaborative managers would benefit from ensuring that expert powers from within are properly used to infuse credibility amongst the transfer, receipt and integration of data-driven knowledge [61].

*7) Sense-making:* Heen [67] explores the performance outcome of satisfactory delivery of primary care medical services and the impact that different managerial roles can have on this success. In her case studies, the municipalities are dependent on co-operation from the regular GPs to solve issues like securing patients, while the GPs' interest is that the network enables them to influence municipal decision in their sector. Thus, the author finds the relationship to be asymmetric: the municipality has more need for cooperation with the GPs than vice versa. The findings from the case studies exemplify that the context of unbalanced reciprocity denotes situations for more game-like activity of indirect management attempting to create strategies for trust building and persuasion. A collaboration manager must, then, make sense of the situational need. For example, fragmentation requires this role to stimulate the formal network structure through enlivening actors to engage themselves. Trust in the diplomacy role is often grounded in authority and tied to the perception that an actor has influence. Next, it is argued that as conflict is inevitable, an adversarial role of management can arise due to the mandatory contractual nature of the network. Because the adversary would openly challenge and confront the opposing parties as someone who is less interested in brokering, and more concerned with maintaining a representative role of the regular administration, distrust is directed towards this figure as a symbol of the municipality, rather than in a personal way. Finally, the partner network manager role indicates that a network has institutionalised, and actors have become integrated, but still requires nurturing. In a similar vein, studies have provided insight into managerial interventions that can be used at each stage in the process of collaboration, highlighting the importance of sense making when choosing which managerial role is needed at each stage of the collaboration [68, 40].

Table 3 summarises the literature on the management of collaborations involving the public sector. The aforementioned seven management strategies for inter-organisational collaboration were elaborated on from the findings in this table.

**TABLE 3** *Management Strategies for Inter-Organisational Collaboration*

| Author(s)/Year | Management Action | Results |
|---|---|---|
| Ansell and Gash (2007), Geedes (2012), Hovik et al. (2015), Klaster et al. (2017), Waugh and Streib (2006) | Designate facilitative leadership at various levels and stages of collaboration (e.g. boundary spanners, 'champions,' during negotiation, etc.) | Conflict-resolution, consensus building, action, inclusive agenda shaping, broad participation, productive group dynamics, empowerment, unity of purpose, and an extended scope |
| Agranoff and McGuire (1999), Ansell and Gash (2017), Chen and Lee (2017) | Promotion of joint action building through creation of shared standards and goals | Develop institutional capacity through less institutional and technical inhibitions |
| Ansell and Gash (2017), Crosby and Bryson (2007), Saz-Carranza and Ospina (2012) | Create learning spaces, a communication strategy, and a compelling vision | Leads to reduced cultural barriers, the development of a sense of commonality amongst stakeholders, and helps to overcome tensions |
| Cuganesan et al (2017), Page (2003), Thomas and Perry (2006) | Induce sharing and stewardship through providing information about skills, resources, policies and examples; make impact of collaborative efforts transparent to create symbols of progress | Employees will change their mind-set in desired way through self-efficacy, certainty, and legitimisation. This stimulates cohesion, innovation and ability to reframe meanings to achieve shared and independent goals |
| Giest (2015), O'Leary and Choi (2012), Chen and Lee (2017), Weber et al. (2008), Agranoff and McGuire (1999) | Use experts to facilitate the demands of highly specialised networks | Expert knowledge helps frame tasks and alternatives ways of conceptualising problems. |
| Heen (2009), Vangen and Winchester (2014), Weber and Khademian (2008) | Understand the situational need of management styles | Adopting practices can help positively control the impact activities have on the diverse culture and power balances |

In sum, the main implications for these strategies include increased conflict-resolution, inclusive agenda shaping, institutional capacity, unity of purpose, and power balance.

# Conclusions

AI holds significant potential to contribute to an ambition to make the UK the world's most innovative economy. In order to fulfil the potential of AI for the UK economy and society, the government is expected to take on an entrepreneurial role in innovation policy through cross-sectoral collaborations with businesses and universities. Examples of such collaborations are already appearing both in the UK and internationally. For example, the Institute for Analytics and Data Science at the University of Essex focuses on delivering AI transformation across the public services in Essex.

AI collaboration may be new, but we have accumulated a lot of experience with similar type ventures in various economic sectors. The potential benefits of bringing public, private, and non-profit actors to collaborate for public service delivery are well known [69]. Nonetheless, these benefits do not come alone. Existing evidence from these types of collaborative ventures suggests that we should be aware of vast managerial complexities and their negative effects on the effectiveness and the value for money of cross-sector collaborations. International evidence suggests that these difficulties do not affect the increasing use of collaborative ventures to deliver public policies across the globe [28]. However, they do contribute to the fact that in several cases, these organisational forms do not achieve the desired results [70].

In order to achieve the highest potential of the AI cross-sectoral collaboration as suggested in the AI Sector Deal, the success factors of similar enterprises around the world must be considered. In a nutshell, we find that facilitative leadership is imperative to collaboration success. In addition, alignment of goals and objectives between all the involved parties has also been identified as a key factor for collaboration success. This is of particular importance when organisations from different sectors collaborate, as managers can utilise how knowledge is represented through the creation of shared standards to promote joint-action for institutional and technical capacity building. A well-defined communication strategy will certainly help to align the interests and expectations of all the members of the collaboration, especially when it comes to discussing the opportunities that data science and AI can bring to a particular project. Socialisation has also been identified as key factor of cross-sectoral collaboration success; which means that, behind all the technical complexities of implementing data science and AI initiatives, policymakers should always transmit the public value that the policy or program ultimately pursues. Moreover, leveraging expert insight will ensure that alternative dimensions to problem solving are incorporated to ensure quality. Lastly, the different micro-management strategies that public managers should use in these ventures would then be contingent to their situational understanding, what has been referred to as sense-making.

# Funding statement

The study is funded by HEFCE Catalyst Fund #E10, and the MINECO CSO2016-80823-P fund.

# References

[1] HM Government. 2017. *Industrial strategy artificial intelligence sector deal*. White Paper.

[2] Mazoni, J. *Civil service transformation*. [Speech] London School of Economics. 24 January 2018.

[3] The Royal Society. 2017. *Machine learning: the power and promise of computers that learn by example* (April 2017): https://royalsociety.org/~/media/policy/projects/machine-learning/ publications/machine-learning-report.pdf. Last accessed 20 May 2018.

[4] Mazzucato, M. 2015. *The entrepreneurial state debunking public vs. private sector myths.* London: Anthem Press.

[5] Colatat, P. 2015. An organizational perspective to funding science: collaborator novelty at DARPA *Research Policy* **44**, 874-887. Available from: https://doi.org/10.1016/j.respol.2015.01.005.

[6] O'Leary R., Bingham, L. 2009. *The collaborative public manager*. Washington, DC: Georgetown Univ. Press.

[7] Mergel, I. 2018. Open innovation in the public sector: drivers and barriers for the adoption of Challenge.gov. *Public Management Review* **20**, 726-745. Available from: https://doi.org/10.1080/14719037.2017.1320044

[8] Eggers, W., Bellman, J. 2015. *The journey to government's digital transformation*. Deloitte. Available from: https://www2.deloitte.com/uk/en/pages/public-sector/articles/the-journey-to-governments-digital-transformation.html. Last accessed 7 February 2018.

[9] Donahue, J.D., Zeckhauser, R.J. 2011. *Collaborative governance: Private roles for public goals in turbulent times*. Princeton: Princeton University Press.

[10] Carty, A. 2012. *How to ensure successful PPP procurement*. European PPP Expertise Centre. Available from: https://www.b2match.eu/system/isc-see2012/files/EIB_Carty_-_AC_Inv_summit_croatia_09MAY12.pdf [Accessed 7 February 2018].

[11] HM Treasury. 2013. *Private Finance Initiative Projects: 2013 Summary Data.* Report number: PU 1587.

[12] Huxham, C. 2003. Theorizing collaboration practice. *Public Management Review* **3**, 401-423. Available from: https://doi.org/10.1080/1471903032000146964

[13] Bovaird, T. 2004. Public–private partnerships: from contested concepts to prevalent practice. *International Review of Administrative Sciences* **70**, 199-215. Available from: https://doi.org/10.1177/0020852304044250

[14] Waugh, W.L., Streib, G. 2006. Collaboration and leadership for effective emergency management. *Public Administration Review* **66**, 131-140. Available from: https://doi.org/10.1111/j.1540-6210.2006.00673.x

[15] Hodge, G.A., Greve, C. 2007. Public-private partnerships: an international performance review. *Public Administration Review* **67**, 545-58. Available from: https://doi.org/10.1111/j.1540-6210.2007.00736.x

[16] McKinsey Global Institute. 2017. *Artificial Intelligence: The next digital frontier?* Report. (http://bit.ly/2E3jReB). Last accessed: 20 May 2018.

[17] Behavioural Insights Team 2017. *Using data science in policy: a report by the Behavioural Insights Team*. Report. (http://bit.ly/2FlL0H3). Last accessed: 20 May 2018.

[18] Viechnicki, P., Eggers, W. 2017. *How much time and money can AI save government?* Report. Deloitte Center for Government Insights. (https://www2.deloitte.com/content/dam/insights/us/articles/3834_How-much-time-and-money-can-AI-save-government/DUP_How-much-time-and-money-can-AI-save-government.pdf. Last accessed: 7 February 2018.

[19] Executive Office of the President National Science and Technology Council. 2016. *Preparing for the future of Artificial Intelligence*. Report. (https://obamawhitehouse.archives.gov/sites/default/files/whitehouse_files/microsites/ostp/NSTC/preparing_for_the_future_of_ai.pdf). Last accessed 4 February 2018.

[20] Yasin, R. 2013. How analytics is making NYC's streets and buildings safer. *GCN Magazine*. (https://gcn.com/articles/2013/10/04/gcn-award-nyc-databridge.aspx). Last accessed: 20 May 2018.

[21] Huxham C., Vangen, S. 2000. Leadership in the shaping and implementation of collaborative agendas: how things happen in a (not quite) joined up world. *Academy of Management Journal*. **43**, 1159-1175.

[22] Riccucci, N. 2010. *Public Administration: Traditions of Inquiry and Philosophies of Knowledge*. Georgetown University Press.

[23] Stoker, G., Smith, G., Maloney, W. 2004. Building social capital in city politics: scope and limitations at interorganisational level. *Political Studies* **52**, 508-530. Available from: https://doi.org/10.1111/j.1467-9248.2004.00493.x

[24] Innes, J., Kaplan, L., Connick, S., Booher, D. 2006. Collaborative governance in the CALFED program: adaptive policy making for California water. *Ecology and Society* **15**, 35.

[25] Innes, J., Connick, S., Booher, D. 2007. Informality as a planning strategy: collaborative water management in the CALFED bay-delta program. *Journal of the American Planning Association* **73**, 195-210. Available from: https://doi.org/10.1080/01944360708976153

[26] Andrews R., Esteve M., T. Ysa. 2015. Public-private joint-ventures: mixing oil and water? *Public Money and Management* **35**, 265-272. Available from: https://doi.org/10.1080/09540962.2015.1047267

[27] Nutt, P. 2006. Comparing public and private sector decision-making practices. *Journal of Public Administration Research and Theory* **16**, 289-318. Available from: https://doi.org/10.1093/jopart/mui041

[28] Hodge, G., Greve, C. 2007. Public-private partnerships: an international performance review. *Public Administration Review* **67,** 545-558. Available from: https://doi.org/10.1111/j.1540-6210.2007.00736.x

[29] House of Lords. 2018. AI in the UK: ready, willing and able?. Select Committee on Artificial Intelligence. Report number: HL Paper 100.

[30] Klijn, E., Teisman, G. 2003. Institutional and strategic barriers to public-private partnerships-an analysis of Dutch cases. *Public Money & Management* **23**, 137-146.

[31] Saz-Carranza, A., Longo, F. 2012. Managing competing institutional logics in public-private joint ventures. *Public Management Review* **14**, 331-357. Available from: https://doi.org/10.1080/14719037.2011.637407

[32] Shaoul, J., Shepherd, A., Stafford, A., Stapleton, P. 2013. *Losing control in joint ventures: The case of building schools for the future*. Report. Institute of Chartered Accountants of Scotland.

[33] Rainey, H. G. 1989. 'Public management—recent research on the political context and managerial roles, structures and behaviors.' *Journal of Management* **15**, 229-250.

[34] Huang, M., Kao, K., Lu, T. and Pu, W. 2009. *Control mechanisms as enhancers of international joint venture performance*. Paper presented at the Academy of Management annual meeting, Chicago.

[35] Perry, J. L., & Wise, L. R. 1990. The motivational bases of public service. *Public Administration Review* **50**, 367-373.

[36] Sonnenberg, F. K. 1992. Partnering: entering the age of co-operation. *Journal of Business Strategy* **13**, 49-52. Available from: https://doi.org/10.1108/eb039494

[37] Healy, P. 2003. Collaborative planning in perspective. *Planning Theory* 2 **(2)**: 101-123. Available from: https://doi.org/10.1177/14730952030022002

[38] Patil, DJ., Munoz, C., Smith, M. 2016. Big Data: A report of algorithmic systems, opportunity, and civil rights. Report. Obama White House Archives. (https://www.hsdl.org/?view&did=792977) Last accessed 9 May 2018.

[39] Chen, Y., & Lee, J. 2018. Collaborative data networks for public service: governance, management, and performance. *Public Management Review* **20**, 672-690. Available from: https://doi.org/10.1080/14719037.2017.1305691

[40] Weber, E., & Khademian, A. 2008. Wicked problems, knowledge challenges, and collaborative capacity builders in network settings. *Public Administration Review* **68**, 334-349. Available from: https://doi.org/10.1111/j.1540-6210.2007.00866.x

[41] Cabrera, A., & Cabrera, E. 2002. Knowledge-sharing dilemmas. *Organization Studies* **23**, 687-710.

[42] Denham, E. *Four lessons NHS Trusts can learn from the oyal Free case*. Information Commissioner's Office blog, 3 July 2017, https://iconewsblog.org.uk/2017/07/03/four-lessons-nhs-trusts-can-learn-from-the-royal-free-case/. Last accessed 20 May 2018.

[43] Holdren, J., & Smith, M. 2016. *Artificial intelligence, automation and the economy*. Report. Executive Office of the President National Science and Technology Council Committee on Technology. (https://obamawhitehouse.archives.gov/sites/whitehouse.gov/files/documents/Artificial-Intelligence-Automation-Economy.PDF) Last accessed 3 February 2018.

[44] Munoz, C., & Smith, M. 2016. 'Big risks, big opportunities: the intersection of big data and civil rights.' Obama White House Archives. https://obamawhitehouse.archives.gov/blog/2016/05/04/big-risks-big-opportunities-intersection-big-data-and-civil-rights. Last accessed: 20 May 2018.

[45] Mandell, M.P. 2001. *Getting results through collaboration: networks and network structures for public policy and management*. Westport: Quorum Books.

[46] McGuire, M. 2006. Collaborative public management: assessing what we know and how we know it. *Public Administration Review* **66**, 33-43. Available from: https://doi.org/10.1111/j.1540-6210.2006.00664.x

[47] Kenis, P., & Provan, K. 2009. Towards an exogenous theory of public network performance. *Public Administration* **87**, 440-456. Available from: 10.1111/j.1467-9299.2009.01775.x

[48] Agranoff, R., & McGuire, M. 1998. Multinetwork management: collaboration and the hollow state in local economic policy. *Journal of Public Administration Research and Theory* **8**, 67-91.

[49] Day, D., Fleenor, J., Atwater, L., Sturm, R., McKee, R. 2014. Advances in leader and leadership development: a review of 25 years of research and theory. *The Leadership Quarterly* **25**, 63-82. Available from: https://doi.org/10.1016/j.leaqua.2013.11.004

[50] Ansell, C., & Gash, A. 2007. Collaborative governance in theory and practice. *Journal of Public Administration Research and Theory*. **18**, 543-571. Available from: https://doi.org/10.1093/jopart/mum032

[51] Hovik, S., & Hanssen, G. 2015. The impact of network management and complexity on multi-level coordination. *Public Administration* **93**, 506-523. Available from: https://doi.org/10.1111/padm.12135

[52] Boyne, G. A., & Chen, A. 2007. Performance targets and public service improvement. *Journal of Public Administration Research and Theory* **17**, 455-477. Available from: https://doi.org/10.1093/jopart/mul007

[53] Madhok, A. 1995. Opportunism and trust in joint venture relationships: an exploratory study and a model. *Scandinavian Journal of Management* **11**, 57-74.

[54] Ethiraj, S. K., & Levinthal, D. 2009. Hoping for A to Z while rewarding only A: complex organizations and multiple goals. *Organization Science* **20**, 4-21.

[55] Jensen, M. C. 2001. Value maximization, stakeholder theory, and the corporate objective function. *Journal of Applied Corporate Finance* **14**, 8-21.

[56] Farnham, D., & Horton, S. 1993. *Managing the new public services*. London: MacMillan.

[57] Copeland, E. Lessons from piloting the London office of data analytics [Lecture] NESTA. 25 May 2017. (http://eddiecopeland.me/lessons-from-piloting-the-london-office-of-data-analytics/)

[58] Ansell, C., & Gash, A. 2017. Collaborative platforms as a governance strategy. *Journal of Public Administration Research and Theory* **28**, 16-32. Available from: https://doi.org/10.1093/jopart/mux030

[59] Saz-Carranza A., & Ospina, S. 2011. The behavioural dimension of governing inter-organizational goal-directed networks: managing the unity-diversity tension. *Journal of Public Administration Research and Theory* **21**, 327-365. Available from: https://doi.org/10.1093/jopart/muq050

[60] Cuganesan, S., Hart, A., Steele, C. 2016. Managing information sharing and stewardship for public-sector collaboration: a management control approach. *Public Management Review* **19**, 862-879. Available from: https://doi.org/10.1080/14719037.2016.1238102

[61] Page, S. 2003. Entrepreneurial strategies for managing interagency collaboration. *Journal of Public Administration Research and Theory* **13**, 311-339.

[62] Podesta, J., Holdren, J., Zients, J., Moniz, E., Pritzker, P. 2014. *Big Data: Seizing Opportunities, Preserving Values*. Report. Obama White House Archives. https://obamawhitehouse.archives.gov/sites/default/files/docs/big_data_privacy_report_may_1_2014.pdf.

[63] Thomson, A., & Perry, J. 2006. Collaboration processes: inside the black box. *Public Administration Review* **66**, 20-32. Available from: https://doi.org/10.1111/j.1540-6210.2006.00663.x

[64] Giest, S. 2015. Network capacity-building in high-tech sectors: opening the black box of cluster facilitation policy. *Public Administration* **93**, 471-489. Available from: https://doi.org/10.1111/padm.12131

[65] Blauer, B. 2017. *Barriers to building a data practice in government* [Lecture] NYU GovLab. (http://sppd.thegovlab.org/lectures/barriers-to-building-a-data-practice-in-government.html) Last accessed 2 February 2018.

[66] Chua, H., Vicente, V., Mintzer, S. 2017. *Looking through the lens of Rijkswaterstaat traffic inspectors and the municipality of Rotterdam*. Blog. Data Science for Social Good-University of Chicago.

(https://dssg.uchicago.edu/2017/08/09/looking-through-the-lens-of-rijkswaterstaat-traffic-inspectors-and-the-municipality-of-rotterdam/) Last accessed 8 February 2018.

[67] Heen, H. 2009. One size does not fit all. *Public Management Review* **11**, 235-253. Available from: https://doi.org/10.1080/14719030802685263

[68] Vangen, S., & Winchester, N. 2014. Managing cultural diversity in collaborations: a focus on management tensions. *Public Management Review* **16**, 686-707. Available from: https://doi.org/10.1080/14719037.2012.743579

[69] Pollit, C., & Bouckaert, G. 2000. *Public management reform: a comparative analysis*. Oxford University Press.

[70] Ysa, T., Sierra, V., Esteve, M. 2014. Determinants of network outcomes: the impact of management strategies. *Public Administration* **92**, 636-655. Available from: https://doi.org/10.1111/padm.12076